\documentclass[conference]{IEEEtran}
\IEEEoverridecommandlockouts
\usepackage{cite}
\usepackage{amsmath,amssymb,amsfonts}
\usepackage{algorithmic}
\usepackage{graphicx}
\usepackage{textcomp}
\usepackage{multirow}

\usepackage{makecell}
\usepackage{pifont}
\newcommand{\cmark}{\ding{51}}%
\newcommand{\xmark}{\ding{55}}%

\def\BibTeX{{\rm B\kern-.05em{\sc i\kern-.025em b}\kern-.08em
    T\kern-.1667em\lower.7ex\hbox{E}\kern-.125emX}}
\begin{document}

\title{Frequency of Interest-based Noise Attenuation Method to Improve Anomaly Detection Performance
\thanks{This work has been submitted to the IEEE for possible publication. Copyright may be transferred without notice, after which this version may no longer be accessible.}
}

\author{\IEEEauthorblockN{1\textsuperscript{st} YeongHyeon Park}
\IEEEauthorblockA{\textit{SK Planet Co., Ltd.} \\
Seongnam, Rep. of Korea \\
yeonghyeon@sk.com}
\and
\IEEEauthorblockN{2\textsuperscript{nd} Myung Jin Kim}
\IEEEauthorblockA{\textit{SK Planet Co., Ltd.} \\
Seongnam, Rep. of Korea \\
myungjin@sk.com}
\and
\IEEEauthorblockN{3\textsuperscript{rd} Won Seok Park}
\IEEEauthorblockA{\textit{SK Planet Co., Ltd.} \\
Seongnam, Rep. of Korea \\
pwswonder@sk.com}
}

\maketitle

\begin{abstract}
Accurately extracting driving events is the way to maximize computational efficiency and anomaly detection performance in the tire frictional nose-based anomaly detection task.
This study proposes a concise and highly useful method for improving the precision of the event extraction that is hindered by extra noise such as wind noise, which is difficult to characterize clearly due to its randomness.
The core of the proposed method is based on the identification of the road friction sound corresponding to the frequency of interest and removing the opposite characteristics with several frequency filters.
Our method enables precision maximization of driving event extraction while improving anomaly detection performance by an average of 8.506\%.
Therefore, we conclude our method is a practical solution suitable for road surface anomaly detection purposes in outdoor edge computing environments.
\end{abstract}


\section{Introduction}
\label{sec:intro}
Anomaly detection by utilizing audio has advantages such as fewer blind spots compared to an image-based anomaly detection system.
However, even in audio, if noise other than the noise of interest such as driving sound is varied and excessive, it interferes to recognize the core information.
In the prior study, we performed an audio-based anomaly detection, which is characterized in an outdoor environment~\cite{park2022road}.
The challenging point of the above task is irregular extra noises such as wind noise that hinder the identification of the friction noise between the tire and the road surface used to determine the abnormality~\cite{park2021noise}.

An effective way to perform surface anomaly detection based on driving noise is to extract only the sound corresponding to the driving event and use it as an input. 
At the same time, as mentioned above, it will be helpful to consider information other than driving noise, which is the disturbance, as noise and attenuate them.
The deep learning-based denoising or noise-robust sound identification method has been recently proposed~\cite{bai2018denoising,rethage2018denoising,ryu2020multisens,lee2020voice}.
However, these methods have limitations in that they have an amount of computation that can be overkill in an edge computing environment, or that the training process takes a lot of cost for dataset preparing~\cite{lee2020cost}.

\begin{figure}[t]
    \begin{center}
        \includegraphics*[width=0.6\linewidth]{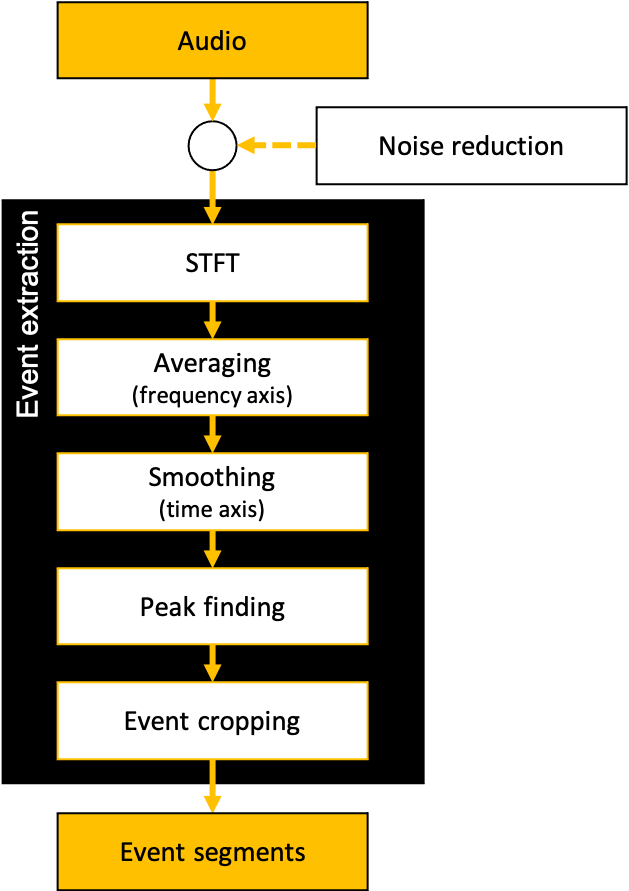} 
	\end{center}
	\vspace*{-4mm}
	\caption{Driving noise extraction minimizes unnecessary computation of the anomaly detection model and helps to make decisions more accurately. Our method includes a noise reduction method to extract driving noise based on the FoI.}
	\label{fig:extraction_flow}
\end{figure}

In this paper, we propose a noise reduction method to improve anomaly detection performance.
More specifically, our goal is to extract only driving noise accurately and to attenuate information that interferes with decision-making for anomaly detection.
Since our purpose is to perform road anomaly detection over a wide area in the outdoor environment while considering edge computing, it is necessary to develop a cost-effective solution by avoiding high-cost methods such as deep learning-based sound event detection~\cite{park2021edge}.
Referring that our method does not require a deep learning model, it does not require the cost and effort for preparing the dataset as prior study~\cite{fu2019event}.

The method presented in this paper is based on the fact that the preprocessing method in the frequency domain is still effective~\cite{ma2022threshold}.
Our band-pass filtering-based noise reduction method is designed to quickly and simply remove the components that interfere with recognizing driving noise.
Through experiments, we show that noise removal is effective to recognize driving sounds with similar shapes and ignore wind noise.
We achieve state-of-the-art precision at event extraction with almost 1.000 value and 0.944 of anomaly detection performance both are improved from before.
For summarizing, we conclude that our noise reduction method is a practical solution with sufficient performance in the real world.

\section{Background}
\label{sec:back}

\subsection{Noise-robust Speech Recognition}
\label{subsec:noise_robust}
Various studies have been conducted for noise-robust speech recognition or sound identification~\cite{bai2018denoising,rethage2018denoising,ryu2020multisens,lee2020voice}.
Some methods that adopt supervised learning~\cite{rethage2018denoising,ryu2020multisens,lee2020voice} are not suitable as approaches for training anomaly detection models that have no choice but to adopt unsupervised methods.

Among the above, a method to improve the recognition performance within the noisy environment can be adopted~\cite{ryu2020multisens,lee2020voice}, but the method of training by synthesizing random noise or random music expands the generalization ability of the anomaly detection model and well reconstruct the data corresponding to the abnormal.
Referring to the anomaly detection system being based on the small reconstruction error of the normal and large error of the abnormal, the above situation will act as a hindrance factor.

We may consider removing the noise from the data first and feeding it into the anomaly detection model.
In this case, there is a limitation that the cost is significantly large because the deep learning-based method is used~\cite{lee2020cost}.

Since we are already using a deep learning-based anomaly detection model, an additional deep learning model for the denoising purpose will increase the cost more.
Thus, instead of the deep learning method, we consider the noise reduction method based on the fact that the frequency domain preprocessing method is still valid~\cite{ma2022threshold}.

\subsection{Wind noise attenuation}
\label{subsec:wind_noise}
Referring to the prior research, it can be confirmed that the frequency characteristics of wind noise are diverse depending on speed and direction~\cite{nemer2009wind, wang2020direction, cook2021wind}.
Although various winds have in common that they occur mainly in the low-frequency band below 3 kHz~\cite{bai2018denoising}, it is difficult to clarify the characteristics~\cite{king2008wind, eldwaik2018wind}.

In addition, it is a known fact that it is difficult to define the characteristics due to the dynamic change caused by turbulence~\cite{peric1997aerody}.
In the case of cities, large differences in speed and direction are observed both horizontal and vertical directions that make it more difficult to define the characteristics of the wind~\cite{ng2011direction,wang2020direction}.

Thus, the data-driven learning approach is adopted in recent~\cite{bai2018denoising,lee2020voice,henry2021review}.
However, as mentioned above, the deep learning-based method is resource-depleting and not appropriate to our purpose.

In this study, we propose a reverse approach based on analyzing the characteristics of the frequency of interest, friction noise between tire and road surface, to attenuate the opposite noise components such as wind noise.

\section{Proposed Method}
\label{sec:propose}

\subsection{Overall view}
\label{subsec:overall}

Via Figure~\ref{fig:extraction_flow}, we present our overall method including noise attenuation based on frequency of interest (FoI).
in succession, we will describe the detail of the two methods, FoI-based noise attenuation and driving event extraction.
The reason for extracting driving event noise for road surface anomaly detection is to minimize unnecessary inference costs of the anomaly detection model at the point in moments when friction noise between the road surface and the tire is not generated.

\subsection{Frequency of Interest-based noise reduction}
\label{subsec:foi_method}
When using the raw audio signal, it may include extra noise other than the friction sound, as same as driving noise or FoI, between the tire and the road surface.
In order to accurately identify the abnormality of the road, developing a noise invariant anomaly detection model or removing the noise should be conducted.
Based on the above need, we focus on reducing the extra noise.

Since the data acquisition device for anomaly detection is in the outdoor environment, it is difficult to attenuate the noise due to the random generation of noise in various frequency bands.
In particular, defining the characteristics of wind noise which is the dominant component of extra noise is challenging due to its variable speed and directions.

Considering that it is challenging to define the characteristics of the extra noise, we propose an approach by identifying the characteristics of the driving noise to attenuate opposite components of the others than FoI.

\begin{figure}[t]
    \begin{center}
        \includegraphics*[width=0.95\linewidth]{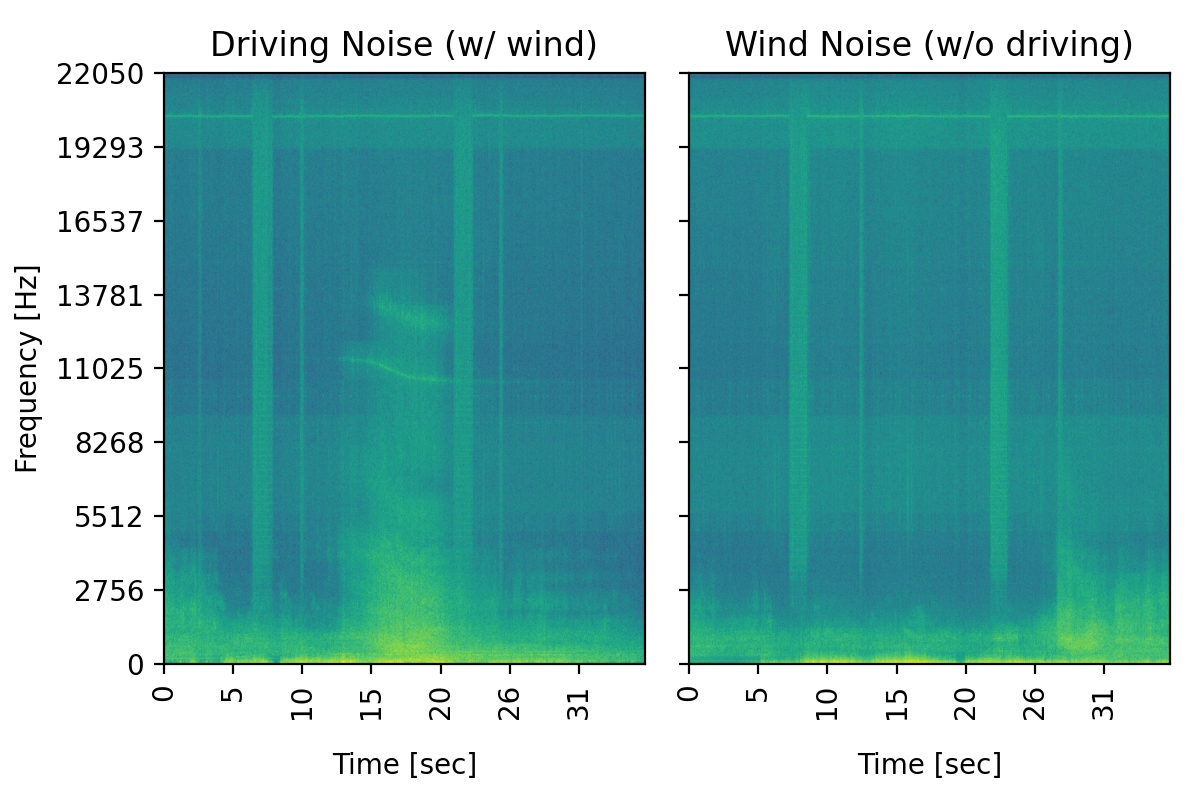} 
	\end{center}
	\vspace*{-4mm}
	\caption{Spectrogram of driving noise and extra (wind) noise. In the case of driving noise (left), the vehicle was passed through the noise acquisition device between 15 and 20 seconds, and the information spans a wide range of frequencies. In contrast, wind noise is represented mainly in the low-frequency band.}
	\label{fig:specgram}
\end{figure}

\begin{figure}[t]
    \begin{center}
        \includegraphics*[width=0.95\linewidth]{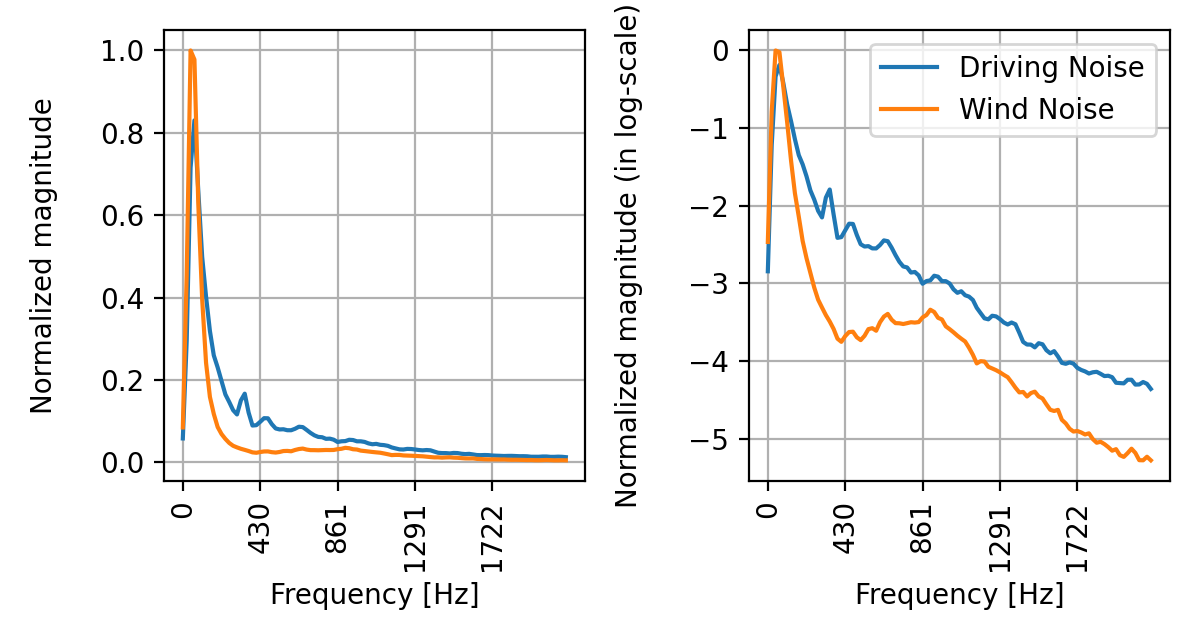}
	\end{center}
	\vspace*{-4mm}
	\caption{The spectrum of low-frequency corresponding to Figure~\ref{fig:specgram}. For visual comparison clearly of the left, we additionally show the log-scaled magnitude on the right side.}
	\label{fig:spectrum}
\end{figure}

Via Figure~\ref{fig:specgram}, we confirm the frequency difference between driving noise and wind noise, which is the extra noise.
Even in the absence of driving noise, a situation may occur in which the accumulated magnitude exceeds a certain level due to strong wind noise.
When using our driving noise extraction method of peak detection based on the cumulative magnitude, the above situation improperly regards the extra noise as driving noise, so we need to attenuate the other noise in advance.

\begin{figure}[t]
    \begin{center}
        \includegraphics*[width=0.95\linewidth]{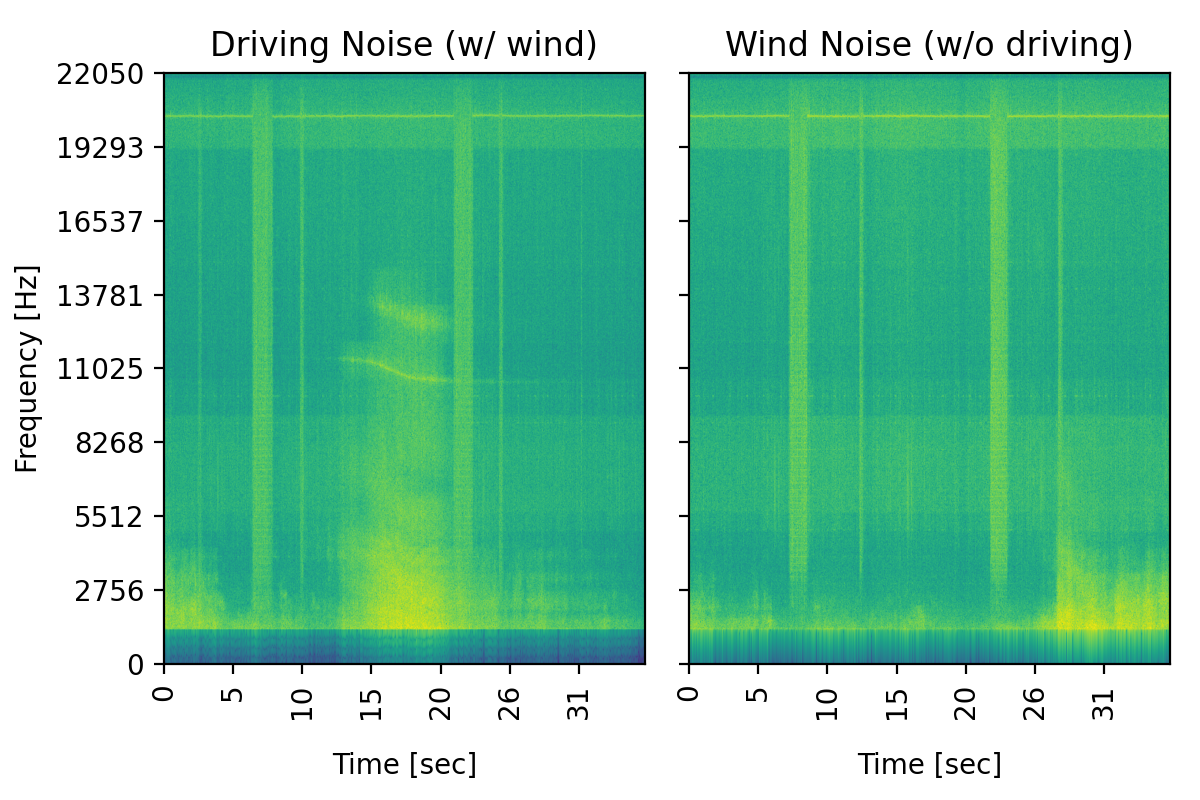}
	\end{center}
	\vspace*{-4mm}
	\caption{Spectrogram after applying noise reduction corresponding to Figure~\ref{fig:specgram}. The low-frequency components are commonly reduced. Meanwhile, on the left side spectrogram, the driving noise that appears between 15 and 20 seconds is not completely reduced as the intention of our method.}
	\label{fig:specgram_reduction}
\end{figure}

\begin{figure}[t]
    \begin{center}
        \includegraphics*[width=0.95\linewidth]{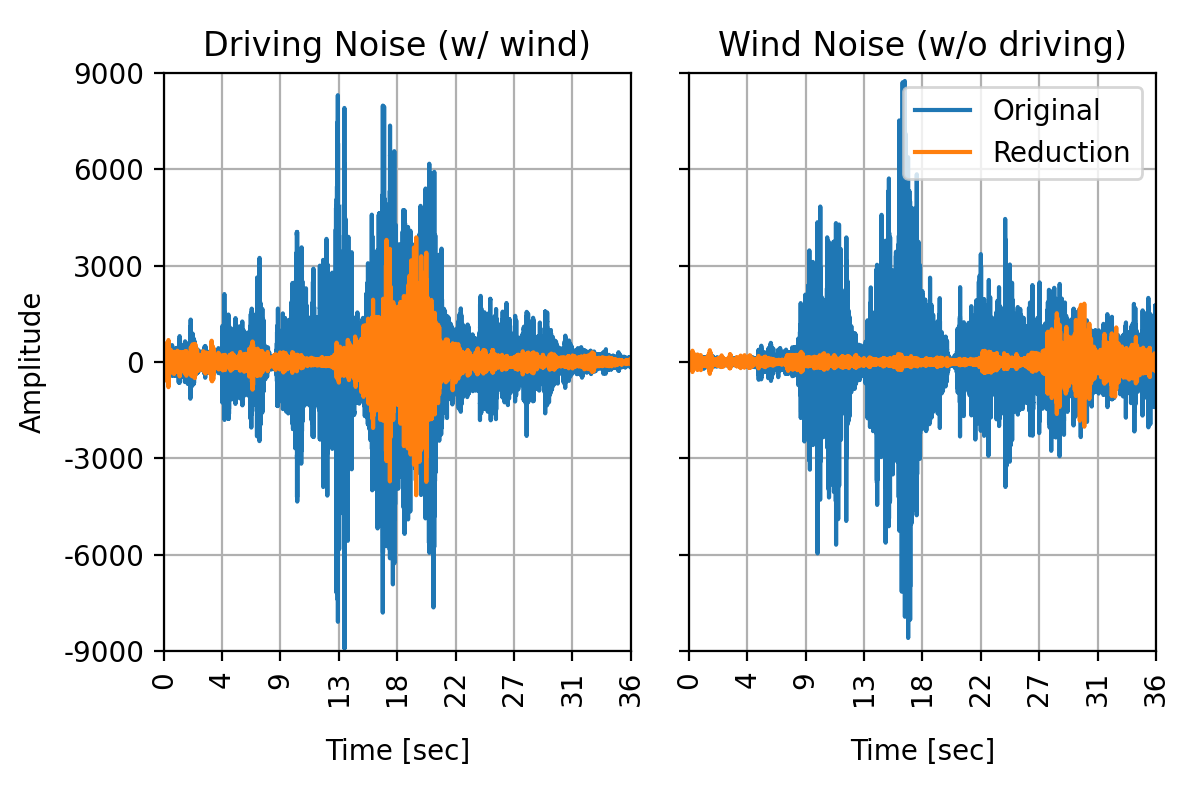}
	\end{center}
	\vspace*{-4mm}
	\caption{Two audio samples before and after noise reduction is shown. Blue and orange colors represent the signal corresponding to Figure~\ref{fig:specgram} and Figure~\ref{fig:specgram_reduction} respectively.}
	\label{fig:noise_reduction}
\end{figure}

Figure~\ref{fig:spectrum} shows the average spectrum through the time-axis to compare the low-frequency information in detail.
Accordingly, we will attenuate low-frequency factors that have a probability to increase the cumulative magnitude on each time step.
When the noise reduction is conducted with a high-pass filter simply, there is a risk that driving information will also be attenuated, so we utilize several notch filters intermittently.
We use 60 harmonics (\textit{1st} to \textit{60th} harmonics) for noise reduction and set the first harmonic as 21.5 Hz and the DC component is also removed.
The result of noise reduction is shown in Figure~\ref{fig:specgram_reduction} and Figure~\ref{fig:noise_reduction}.

\subsection{Peak detection-based event extraction}
\label{subsec:foi_method}
Event extraction method is already shown in Figure~\ref{fig:extraction_flow}.
We first perform noise reduction and short-time Fourier transform (STFT) to construct frequency information for each time step.
Then, we calculate the average magnitude on the frequency axis in order to minimize the cost of peak detection or event detection.
Note that, we are not adopting the deep learning-based sound event detection algorithm to avoid the delay in the anomaly detection process and computational burden on edge device.

To merge the events of two or more vehicles passing at almost the same moment, we perform the smoothing through the hanning windowing.
Finally, we utilize the peak detection algorithm provided in Scipy to find the center of the driving event and crop them with a 5-second margin back and forth (10 seconds length).
We feed this into the anomaly detection model to determine whether there are any abnormalities.

\section{Experiments}
\label{sec:exps}

\subsection{Dataset}
\label{subsec:dataset}
We collected the dataset in three different locations with four weather conditions.
Each audio sample is recorded in a 10-minute length with sampling rate of 44.1kHz.
Basically, our problem is to treat and detect foreign substances including black ice on the road surface, so we simply label the data from the dry condition as normal and other conditions as abnormal.
In order to properly perform anomaly detection, the precision of driving event extraction is more important than the amount of extracted driving events, therefore other metrics are not covered in this study.

\begin{table}[t]
    \centering
    \caption{Dataset for experiments acquired from the three different locations. Each sample is recorded in a 10-minute length.}
    \begin{tabular}{l|r|rrr}
        \hline
            \multirow{2}{*}{\textbf{Post}} & \textbf{Normal} & \multicolumn{3}{c}{\textbf{Abnormal}} \\
            \cline{2-5}
             & \textbf{Dry} & \textbf{Wet} & \textbf{Slush} & \textbf{Snow} \\
        \hline
        \hline
            A & 10 & 10 & 4 & - \\
        \hline
            B & 10 & 10 & 2 & - \\
        \hline
            C & 10 & 9 & 10 & 3\\
        \hline
        \hline
            Total & 30 & 29 & 16 & 3 \\
        \hline
    \end{tabular}
    \label{tab:dataset}
\end{table}

\subsection{Improvement of event extraction}
\label{subsec:exp_event}
First, we measure the precision of event extraction with or without noise reduction methods applied.
Table~\ref{tab:extracted_event} summarizes the precision according to the condition by displaying each location as a post.

\begin{table}[t]
    \centering
    \caption{Comparison of event extraction precision. For each case, the ratio of driving events among the extracted events is displayed as 'number of driving events/amount of extracted events', and the precision from the above is also presented.}
    \resizebox{\columnwidth}{!}{%
    \begin{tabular}{l|c|r|rrr}
        \hline
            \multirow{2}{*}{\textbf{Post}} & \multirow{2}{*}{\textbf{Reduction}} & \textbf{Normal} & \multicolumn{3}{c}{\textbf{Abnormal}} \\
            \cline{3-6}
             & & \textbf{Dry} & \textbf{Wet} & \textbf{Slush} & \textbf{Snow} \\
        \hline
        \hline
            \multirow{4}{*}{A} 
            & \multirow{2}{*}{\xmark} & 513 / 570 & 41 / 70 & 10 / 13 & - \\
            & & (0.900) & (0.586) & (0.769)- & \\
            \cline{2-6}
            & \multirow{2}{*}{\cmark} & 384 / 385 & 21 / 21 & 7 / 7 & - \\
            & & (\textbf{0.966}) & (\textbf{1.000}) & (\textbf{1.000})- & \\
        \hline
        \hline
            \multirow{4}{*}{B} 
            & \multirow{2}{*}{\xmark} & 1492 / 1545 & 568 / 568 & 163 / 253 & - \\
            & & (0.966) & (\textbf{1.000}) & (0.644)- & \\
            \cline{2-6}
            & \multirow{2}{*}{\cmark} & 804 / 804 & 529 / 529 & 11 / 11 & - \\
            & & (\textbf{1.000}) & (\textbf{1.000}) & (\textbf{1.000})- & \\
        \hline
        \hline
            \multirow{4}{*}{C} 
            & \multirow{2}{*}{\xmark} & 1336 / 1337 & 1013 / 1013 & 149 / 246 & 15 / 55 \\
            & & (0.999) & (\textbf{1.000}) & (0.606) & (0.273) \\
            \cline{2-6}
            & \multirow{2}{*}{\cmark} & 1153 / 1153 & 1032 / 1032 & 76 / 76 & 4 / 4 \\
            & & (\textbf{1.000}) & (\textbf{1.000}) & (\textbf{1.000}) & (\textbf{1.000}) \\
        \hline
    \end{tabular}
    }
    \label{tab:extracted_event}
\end{table}

We confirm that when the noise reduction method is applied the amount of event extraction is slightly reduced but the precision is significantly improved.
Moreover, since we achieve precision 1 in our dataset, we guarantee that the anomaly detection model can always take noise corresponding to the driving events.

\subsection{Improvement of anomaly detection}
\label{subsec:exp_event}
Anomaly detection experiments are performed based on the extracted events by applying noise reduction.
The number of the extracted event for each condition is summarized in Table~\ref{tab:dataset}.
Event extraction is splitting equal points in the raw audio (including noise) utilizing selected peaks of our algorithm basically.
In this section, we perform an experiment to confirm if the anomaly detection performance can be further improved by applying noise reduction once more on the extracted events.
This is equivalent to extracting the event as it is from the noise-reduced audio.

\begin{table}[t]
    \centering
    \caption{Extracted driving events for anomaly detection experiments}
    \begin{tabular}{l|rr}
        \hline
            \textbf{Post} & \textbf{Normal} & \textbf{Abnormal} \\
        \hline
        \hline
            A & 384 & 28 \\
        \hline
            B & 804 & 540 \\
        \hline
            C & 1153 & 1112 \\
        \hline
    \end{tabular}
    \label{tab:dataset}
\end{table}

We summarize the the receiver operating characteristic curve (AUROC)~\cite{fawcett2006auroc} in Table~\ref{tab:performance_ad} to compare tue anomaly detection performance.
We utilize the simple autoencoder-based anomaly detection model non-compressed auto-encoder (NCAE)~\cite{park2022road}.
As a result, we achieve improvement in anomaly detection performance using noise reduction of at least 4.062\% and up to 12.360\% for each independent post.
In particular, the experiment that merges the three posts (marked as Post-merge) shows a dramatic improvement of 39.908\%.

\begin{table}[t]
    \centering
    \caption{Comparison of anomaly detection performance}
    \begin{tabular}{l|rr|r}
        \hline
            \textbf{Post} & \textbf{Original} & \textbf{Reduction} & \textbf{Improvement} \\
        \hline
        \hline
            A & 0.883 & \textbf{0.963} & 9.060\% \\
        \hline
            B & 0.837 & \textbf{0.871} & 4.062\% \\
        \hline
            C & 0.890 & \textbf{1.000} & 12.360\% \\
        \hline
        \hline
            Average & 0.870 & \textbf{0.944} & 8.506\% \\
        \hline
        \hline
            Post-Merge & 0.654 & \textbf{0.915} & 39.908\% \\
        \hline
    \end{tabular}
    \label{tab:performance_ad}
\end{table}

\section{Conclusion}
\label{sec:conclusion}
In this study, we propose a method to improve anomaly detection performance and robustness to various noises other than the FoI encountered in the outdoor environment.
Considering the characteristics of the road surface anomaly detection task that may have to consider the resource-constrained edge computing environment, we focus on developing a concise and powerful method as much as possible to minimize the computational load other than the anomaly detection model.

Our method actively utilizes well-known Fourier transform and peak detection methods while also including some tricks to minimize the loss of important information.
As shown results, our method not only maximizes the precision of driving event extraction but also improves anomaly detection performance further.
In conclusion, we confirm that our method is a practical solution in the real world.

\section*{Acknowledgements}
We are grateful to all the members of SK Planet Co., Ltd., who have supported this research, providing equipment for the experiment.

\end{document}